\newif\ifcomment\commentfalse

\documentclass[11pt,a4paper]{article}
\usepackage[hyperref]{style/emnlp-ijcnlp-2019}
\usepackage{times}
\usepackage{graphicx}
\usepackage{subfigure} 
\usepackage{adjustbox}
\usepackage{rotating}

\usepackage{natbib}
\usepackage{algorithm}
\usepackage{algorithmic}
\usepackage{multirow}

\usepackage{hyperref}
\usepackage{amsmath}
\usepackage{amsfonts}
\usepackage{amssymb}
\usepackage{arydshln}
\usepackage{bm}
\usepackage{ifthen}
\usepackage{xcolor}
\usepackage{booktabs}
\usepackage[utf8]{inputenc}
\usepackage{url}
\usepackage[misc]{ifsym}

\ifthenelse{\isundefined{\definition}}{\newtheorem{definition}{Definition}}{}
\ifthenelse{\isundefined{\assumption}}{\newtheorem{assumption}{Assumption}}{}
\ifthenelse{\isundefined{\hypothesis}}{\newtheorem{hypothesis}{Hypothesis}}{}
\ifthenelse{\isundefined{\proposition}}{\newtheorem{proposition}{Proposition}}{}
\ifthenelse{\isundefined{\theorem}}{\newtheorem{theorem}{Theorem}}{}
\ifthenelse{\isundefined{\lemma}}{\newtheorem{lemma}{Lemma}}{}
\ifthenelse{\isundefined{\corollary}}{\newtheorem{corollary}{Corollary}}{}
\ifthenelse{\isundefined{\alg}}{\newtheorem{alg}{Algorithm}}{}
\ifthenelse{\isundefined{\example}}{\newtheorem{example}{Example}}{}
\aclfinalcopy % Uncomment this line for the final submission
\title{\textsc{XGLUE}: A New Benchmark Dataset \\for Cross-lingual  Pre-training, Understanding and Generation}

\author { \space \\
Yaobo Liang, {\Letter Nan Duan}, Yeyun Gong, Ning Wu, Fenfei Guo, Weizhen Qi, Ming Gong, Linjun Shou, \\ Daxin Jiang, Guihong Cao, Xiaodong Fan, Ruofei Zhang, Rahul Agrawal, Edward Cui, Sining Wei, Taroon Bharti, \\ Ying Qiao, Jiun-Hung Chen, Winnie Wu, Shuguang Liu, Fan Yang, Daniel Campos, Rangan Majumder, Ming Zhou \\
\{yalia,nanduan,yegong,t-niwu,v-fengu,v-weqi,migon,lisho,djiang,gucao,xiafan,bzhang,rahul.agrawal,edwac,\\sinwei,tbharti,yiqia,jiuche,winniew,shuguanl,fanyang,Campos.Daniel, ranganm,mingzhou\}@microsoft.com
}

\begin{document}
\maketitle

\begin{abstract}
In this paper, we introduce \textbf{XGLUE}, a new benchmark dataset that can be used to train large-scale cross-lingual pre-trained models using multilingual and bilingual corpora and evaluate their performance across a diverse set of cross-lingual tasks. Comparing to GLUE \cite{glue}, which is labeled in English for natural language understanding tasks only, XGLUE has two main advantages: 
%(1) it provides two corpora with different sizes for cross-lingual pre-training; 
(1) it provides 11 diversified tasks that cover both natural language understanding and generation scenarios; 
(2) for each task, it provides labeled data in multiple languages. We extend a recent cross-lingual pre-trained model Unicoder \cite{huang2019unicoder} to cover both understanding and generation tasks, which is evaluated on XGLUE as a strong baseline. We also evaluate the base versions (12-layer) of Multilingual BERT, XLM and XLM-R for comparison.
\end{abstract}

\section{Introduction}
\label{sec:intro}

Pre-training + Fine-tuning has become a new NLP paradigm, where the general knowledge are firstly learnt from large-scale corpus by self-supervised learning and then transferred to downstream tasks by task-specific fine-tuning. Three different types of pre-trained models are explored recently, including \textit{monolingual pre-trained models} \cite{gpt,devlin-18,liu2019roberta,xlnet,unilm,lewis2019bart}, \textit{multilingual and cross-lingual pre-trained models} \cite{devlin-18,conneau2019cross,huang2019unicoder,xlmr} and \textit{multimodal pre-trained models} \cite{vilbert,unicodervl,uniter,vlp}.
In this paper, we focus on the cross-lingual pre-trained models, due to their importance to alleviating the low-resource issue among languages, where an NLP task often has rich training data in one language (such as English) but has few or no training data in other languages (such as French and German).
In order to further advance the development of cross-lingual pre-trained models for various downstream tasks in different languages, this paper introduces \textbf{XGLUE}, a new benchmark dataset that can be used to: (i) train large-scale  cross-lingual pre-trained models using multilingual and bilingual corpora, (ii) evaluate generalization capabilities of the cross-lingual pre-trained models across a diverse set of cross-lingual tasks.

The contribution of XGLUE is two-fold. 
First, it provides 11 diversified cross-lingual tasks covering both understanding and generation scenarios. 
%which, to the best of our knowledge, is the first attempt in the cross-lingual dataset construction efforts. 
XTREME \cite{XTREME} is a concurrent work of XGLUE. But it includes cross-lingual understanding tasks only. Besides, XGLUE introduces 6 new tasks selected from Search, Ads and News scenarios,which makes XGLUE have more practical values.
Second, an extended version of Unicoder \cite{huang2019unicoder} is described and evaluated as a strong cross-lingual pre-trained model baseline on XGLUE for both understanding and generation tasks. 
We also evaluate the base versions (12-layer) of Multilingual BERT \cite{devlin-18}, XLM \cite{conneau2019cross} and XLM-R \cite{xlmr} for comparison.
%We conduct comprehensive experiments on XGLUE, which not only show interesting findings, but also point out several ways to further improve the cross-lingual pre-trained models.
\section{XGLUE Benchmark\footnote{https://microsoft.github.io/XGLUE/}}
\subsection{Pre-training Corpus}

We collect two corpora, \textit{Small Corpus} and \textit{Large Corpus}, with different sizes for cross-lingual pre-training. 
%the former can be used to evaluate new ideas effectively and the latter can be used to train large-scale models.
Table 1 lists the data statistics.

%Our pre-training corpus contains multilingual dataset and Bilingual dataset from multiple languages. For multilingual dataset, we introduced two datasets, Wikipedia and Common Crawl. We suggest to compare the model performance on same dataset. For bilingual dataset, we introduce a collection of open MT corpus.
\subsubsection{Small Corpus (SC)}
\paragraph{Multilingual Corpus}
We extract raw sentences from Wikipedia using \emph{WikiExtractor}. 
It leads to a 101G multilingual corpus covering 100 languages.

\paragraph{Bilingual Corpus}
We use an in-house pipeline to extract bilingual sentence pairs from the Web, which leads to a 146G bilingual corpus covering 27 languages, including \emph{Arabic}, \emph{Bulgarian}, \emph{Danish}, \emph{German}, \emph{Greek}, \emph{English}, \emph{Spanish}, \emph{Finnish}, \emph{French}, \emph{Hebrew}, \emph{Hindi}, \emph{Hungarian}, \emph{Indonesian}, \emph{Italian}, \emph{Japanese}, \emph{Korean}, \emph{Dutch}, \emph{Polish}, \emph{Portuguese}, \emph{Russian}, \emph{Swedish}, \emph{Swahili}, \emph{Thai}, \emph{Turkish}, \emph{Urdu}, \emph{Vietnamese} and \emph{Chinese}.

\subsubsection{Large Corpus (LC)}
\paragraph{Multilingual Corpus}
Following \citet{wenzek2019ccnet}, we construct a clean version of Common Crawl (CC)\footnote{https://commoncrawl.org/.} as the multilingual corpus. First, we use a language identification model trained based on Wikipedia to classify the language of each page in CC. Then, we train a language model for each language using the corresponding part of the Wikipedia corpus, and use it to filter documents as \citet{wenzek2019ccnet} did. We use one CC dump for English and twelve CC dumps for other languages. It leads to a 2,500G multilingual corpus covering 89 languages. We also include the 101G multilingual corpus described in Section 2.1.1.

\paragraph{Bilingual Corpus}
We reuse the bilingual corpus described in Section 2.1.1. We will add CCMatrix \cite{ccmatrix} in the future.

\begin{table}
  \small
	\centering
		\resizebox{1\linewidth}{!}{
	\begin{tabular}{lcccc}
		\toprule
		 & Type & \# of Languages & Size \\
		\midrule
		\midrule
		\multirow{2}{*}{Small Corpus} & Multilingual & 100 & 101G \\
		 & Bilingual & 27 & 146G \\
		\midrule
		\multirow{2}{*}{Large Corpus} & Multilingual & 100 & 2,500G+101G \\
         & Bilingual & 27 & 146G \\
		\bottomrule
	\end{tabular}
	}
	\caption{The statistics of two pre-training corpora.}
	\label{tab:pre-train-dataset} 
\end{table}

\subsection{Downstream Tasks}
\label{sec:tasks}

\begin{table*}[t]
  \small
	\centering
    \resizebox{1\linewidth}{!}{
	\begin{tabular}{lcccccccc}
		\toprule
		Task & \# of Languages & $\vert$Train$\vert^{en}$ & $\vert$Dev$\vert^{avg}$ & $\vert$Test$\vert^{avg}$ & Metric & Data Source\\
		\midrule
		\midrule 
		\multicolumn{1}{l}{NER}         & 4 &  15.0K & 2.8K & 3.4K & F1 & ECI Multilingual Text Corpus \\
        \multicolumn{1}{l}{POS}         & 18 & 25.4K & 1.0K & 0.9K & ACC & UD Tree-banks~(v2.5)  \\
        \multicolumn{1}{l}{NC$^*$}      & 5 & 100K & 10K & 10K & ACC & Commercial News Website \\
        \multicolumn{1}{l}{MLQA}        & 7 & 87.6K & 0.6K & 5.7K & F1 & Wikipedia  \\
        \multicolumn{1}{l}{XNLI}        & 15 & 433K & 2.5K & 5K & ACC & MultiNLI Corpus \\
        \multicolumn{1}{l}{PAWS-X}      & 4 & 49.4K & 2K & 2K & ACC & Wikipedia \\
        \multicolumn{1}{l}{QADSM$^*$}   & 3 & 100K & 10K & 10K & ACC & Commercial Search Engine \\
        \multicolumn{1}{l}{WPR$^*$}     & 7 & 100K & 10K & 10K & nDCG & Commercial Search Engine \\
        \multicolumn{1}{l}{QAM$^*$}     & 3 & 100K & 10K & 10K & ACC & Commercial Search Engine \\
        \multicolumn{1}{l}{QG$^*$}      & 6 & 100K & 10K & 10K & BLEU-4 & Commercial Search Engine  \\
        \multicolumn{1}{l}{NTG$^*$}     & 5 & 300K & 10K & 10K & BLEU-4 & Commercial News Website  \\
		\bottomrule
	\end{tabular}
	}
	\caption{11 downstream tasks in XGLUE. For each task, training set is only available in English. $\vert$Train$\vert^{en}$ denotes the number of labeled instances in the training set. $\vert$Dev$\vert^{avg}$ and $\vert$Test$\vert^{avg}$ denote
	the average numbers of labeled instances in the dev sets and test sets, respectively. $*$ denotes the corresponding dataset is constructed by this paper.}
	\label{tab:statistics}
\end{table*}

\begin{table*}[t]
  \small
	\centering
	\resizebox{1\linewidth}{!}{
	\begin{tabular}{lccccccccccccccccccc}
		\toprule
		Task & ar & bg & de &el & en & es & fr & hi & it & nl & pl & pt & ru & sw & th & tr & ur & vi & zh \\
		\midrule
		\midrule 
		\multicolumn{1}{l}{NER} &  &  & \checkmark & & \checkmark & \checkmark &  &  &  & \checkmark &  &  &  &  &  &  &  &  &   \\
		\multicolumn{1}{l}{POS} & \checkmark & \checkmark & \checkmark & \checkmark & \checkmark & \checkmark & \checkmark & \checkmark & \checkmark & \checkmark & \checkmark & \checkmark & \checkmark &  & \checkmark & \checkmark & \checkmark & \checkmark & \checkmark \\
		\multicolumn{1}{l}{NC$^*$} &  &  & \checkmark & & \checkmark & \checkmark & \checkmark &  & &  &  & & \checkmark &  &  &  &  &  &  \\
        \multicolumn{1}{l}{MLQA} & \checkmark &  & \checkmark & & \checkmark & \checkmark &  & \checkmark &  &  &  &  &  &  &  &  &  & \checkmark & \checkmark \\
        \multicolumn{1}{l}{XNLI} & \checkmark & \checkmark & \checkmark &\checkmark & \checkmark & \checkmark & \checkmark & \checkmark &  &  &  &  & \checkmark & \checkmark & \checkmark & \checkmark & \checkmark & \checkmark & \checkmark \\
        \multicolumn{1}{l}{PAWS-X} &  &  & \checkmark & & \checkmark & \checkmark & \checkmark &  &   &  &  &  &  &  &  &  &  &  &  \\
        \multicolumn{1}{l}{QADSM$^*$} &  &  & \checkmark & & \checkmark &  & \checkmark &  &  &  &  &  &  &  &  &  &  &  &   \\
        \multicolumn{1}{l}{WPR$^*$} &  &  & \checkmark & & \checkmark & \checkmark & \checkmark &  & \checkmark &  &  & \checkmark &  &  &  &  &  &  & \checkmark \\
        \multicolumn{1}{l}{QAM$^*$} &  &  & \checkmark & & \checkmark &  & \checkmark &  &  &  &  &  &  &  &  &  &  &  &  \\
        \multicolumn{1}{l}{QG$^*$} & &  & \checkmark  & & \checkmark & \checkmark & \checkmark &  & \checkmark &  & &\checkmark & &  &  &  &  & &  \\ 
        \multicolumn{1}{l}{NTG$^*$} &  &  & \checkmark & & \checkmark & \checkmark & \checkmark &  &  &  &  & & \checkmark &  &  &  &  &  &  \\

		\bottomrule
	\end{tabular}
	}
	\caption{The 19 languages covered by the 11 downstream tasks: \textit{Arabic} (ar), \textit{Bulgarian} (bg), \textit{German} (de), \textit{Greek} (el), \textit{English} (en), \textit{Spanish} (es), \textit{French} (fr), \textit{Hindi} (hi), \textit{Italian} (it), \textit{Dutch} (nl), \textit{Polish} (pl), \textit{Portuguese} (pt), \textit{Russian} (ru), \textit{Swahili} (sw), \textit{Thai} (th), \textit{Turkish} (tr), \textit{Urdu} (ur), \textit{Vietnamese} (vi), and \textit{Chinese} (zh).
	All these 6 new tasks with $*$ are labeled by human, except es, it and pt datasets in QG (80+\% accuracy) are obtained by an in-house QA ranker.}
	\label{tab:task-language} 
\end{table*}

We select 11 cross-lingual tasks in XGLUE, which are categorized into 3 groups: single-input understanding tasks, pair-input understanding tasks, and generation tasks. For each task, training set is only available in English. In order to obtain a good performance on XGLUE, a model should be able to learn how to do a task well using its English training set, and then transfer this ability to test sets in other languages.
Table 2 gives the dataset statistics and Table 3 lists languages covered by all tasks.

\subsubsection{Single-input Understanding Tasks}

\paragraph{NER}
We select a subset of the following two NER tasks, CoNLL-2002 NER \cite{sang2002ef} and CoNLL-2003 NER \cite{sang2003introduction}, to form this cross-lingual NER dataset. 
It covers 4 languages, including \emph{English}, \emph{German}, \emph{Spanish} and \emph{Dutch}, and 4 types of named entities, including \emph{Person}, \emph{Location}, \emph{Organization} and \emph{Miscellaneous} entities that do not belong to the previous three types. F1 score is used as the metric.

\paragraph{POS Tagging (POS)}
Following \cite{kim-etal-2017-cross}, we select a subset of Universal Dependencies (UD) Treebanks (v2.5) \cite{ud2.5}, which covers 18 languages. Accuracy (ACC) of the predicted POS tags is used as the metric.

\paragraph{News Classification (NC)}
This task aims to predict the category given a news article.  It covers 5 languages, including \emph{English}, \emph{Spanish}, \emph{French}, \emph{German} and \emph{Russian}. 
Each labeled instance is a 3-tuple: $<$news title, news body, category$>$.
%The category set includes foodanddrink", "sports", "news", "entertainment", "health", "video", "finance", "travel", "lifestyle", "autos.
The category number is 10.
We crawl this dataset from a commercial news website.
Accuracy (ACC) of the multi-class classification is used as the metric.

\subsubsection{Pair-input Understanding Tasks}

\paragraph{MLQA}
The MLQA \cite{lewis2019mlqa} is a multilingual machine reading comprehension task, which contains QA annotations labeled in 7 languages, including \emph{English},  \emph{Arabic}, \emph{German}, \emph{Spanish}, \emph{Hindi}, \emph{Vietnamese} and \emph{Chinese}. 
%It consists of 12K QA instances in English and 5K in each other language, with each instance being parallel between 4 language on average. 
F1 score of the predicted answers is used as the metric.

\paragraph{XNLI}
We reuse the original XNLI dataset \cite{conneau2018xnli} in XGLUE.

\paragraph{PAWS-X}
The PAWS-X \cite{yang2019paws} is a paraphrase identification dataset, which extends the Wikipedia portion of the PAWS \cite{zhang2019paws} evaluation to more languages. We select 4 languages, including \emph{English}, \emph{Spanish}, \emph{French} and \emph{German}, from the original dataset and use them in XGLUE. Accuracy (ACC) of the binary classification is used as the metric.

\paragraph{Query-Ad Matching (QADSM)}
This task aims to predict whether an advertisement (ad) is relevant to an input query. 
It covers 3 languages, including \emph{English}, \emph{French} and \emph{German}. 
Each labeled instance is a 4-tuple: $<$query, ad title, ad description, label$>$.
The label indicates whether the ad is relevant to the query (Good), or not (Bad). 
We construct this dataset based on a commercial search engine.
Accuracy (ACC) of the binary classification is used as the metric.

\paragraph{Web Page Ranking (WPR)}
This task aims to predict whether a web page is relevant to an input query. 
It covers 7 languages, including \emph{English}, \emph{German}, \emph{French}, \emph{Spanish}, \emph{Italian}, \emph{Portuguese} and \emph{Chinese}. 
Each labeled instance is a 4-tuple: $<$query, web page title, web page snippet, label$>$. 
The relevance label contains 5 ratings: Perfect (4), Excellent (3), Good (2), Fair (1) and Bad (0). 
We construct this dataset based on a commercial search engine. 
Normalize Discounted Cumulative Gain (nDCG) is used as the metric.

\paragraph{QA Matching (QAM)}
This task aims to predict whether a $<$question, passage$>$ pair is a QA pair.  It covers 3 languages, including \emph{English}, \emph{French} and \emph{German}. Each labeled instance is a 3-tuple: $<$question, passage, label$>$. The label indicates whether the passage is the answer of the question (1), or not (0). 
We construct this dataset based on a commercial search engine.
Accuracy (ACC) of the binary classification is used as the metric.

\subsubsection{Generation Tasks}

\paragraph{Question Generation (QG)}
This task aims to generate a question for a given passage. 
We collect $<$passage, question$>$ pairs from a commercial search engine. 
It covers 6 languages, including \emph{English}, \emph{French}, \emph{German}, \emph{Spanish}, \emph{Italian} and \emph{Portuguese}. 
BLEU-4 score is used as the metric.

\paragraph{News Title Generation (NTG)}
This task aims to generate a proper title for a given news body. 
We collect $<$news body, news title$>$ pairs from a commercial news website.
It covers 5 languages, including \emph{German}, \emph{English}, \emph{French}, \emph{Spanish} and \emph{Russian}. 
BLEU-4 score is used as the metric.
\section{Pre-train Unicoder for Cross-lingual Understanding Tasks}
We select Unicoder \cite{huang2019unicoder} as the backbone model. 
Section 3 introduces a simplified version of Unicoder using two pre-training tasks (MLN and TLM) for cross-lingual understanding tasks.
Section 4 describes how to extend Unicoder to cover cross-lingual generation tasks.  

The original Unicoder \cite{huang2019unicoder} includes more pre-training tasks besides MLM and TLM. But to keep the baseline pre-trained model simple and to reduce the experimental cost, we just use MLM and TLM in this paper. 
It means for understanding tasks, Unicoder is almost equal to XLM, except some hyper-parameter differences.
%\textit{We will add the results of Unicoder pre-trained by more tasks beyond MLM and TLM in the updated version.}

\subsection{Masked Language Model (MLM)}
Following \citet{devlin-18}, this task extends the masked language model task to multiple languages.
At each iteration, a batch is composed of sentences sampled from different languages.  The sampling probability of a language $l_i$ is defined as $\lambda_{l_i}= p_{l_i}^{\alpha}/\sum_{l_i} p_{l_i}^{\alpha}$,
%\begin{equation}
%   \lambda_{l_i}= \frac{p_{l_i}^{\alpha}}{\sum_{l_i} p_{l_i}^{\alpha}}  \nonumber
%\end{equation}
where $p_{l_i}$ is the percentage of the language $l_i$ in the entire corpus, the smoothing factor $\alpha$ is set to 0.3.
For each batch, we randomly sample 15\% of the words and replace them with (i) a special symbol [MASK], (ii) a random token or (iii) keep them unchanged with probability 80\%, 10\% and 10\%, respectively.
For each token, we only use its token embedding and position embedding, and discard segment embedding and language embedding.

\subsection{Translation Language Model (TLM)}
Following \citet{conneau2019cross}, this task extends the MLM task to bilingual corpus. Given a bilingual sentence pair, TLM first concatenates them into a single sentence, and then masks words using the same strategy of MLM. The pre-trained model learns to recover each masked word based on the bilingual context. We follow MLM to sample language pairs in each batch with $\alpha=0.3$.

\section{Pre-train Unicoder for Cross-lingual Generation Tasks}
\label{sec:model}

The encoder-decoder architecture is employed to extend Unicoder to generation tasks, where the BPE embeddings are shared between encoder and decoder.
%, which is initialized based on the well-trained Unicoder with shared BPE embeddings. 
Two separate generative tasks are proposed for Unicoder pre-training: \textit{Multilingual Denoising Auto-Encoding (xDAE)} and \textit{Multilingual Future N-gram Prediction (xFNP)}.

\subsection{Multilingual Denoising Auto-Encoding (xDAE)}
Motivated by BART \cite{lewis2019bart}, xDAE aims to predict the original text $X = (x_1,x_2,...,x_{|X|}) \in l_i$ from a language $l_i$ based on its corrupted form $c(X)$, where $c(X)$ is a noising function that corrupts an input text $X$ as its output. 

Four different text noising strategies for $c(\cdot)$ are explored in this paper. (1) Shuffle the input text $X$ by adding a noise $\alpha \sim {\rm U}(0, 3)$ to the input indices and then re-ordering $X$ based on the rank of the noised indices. (2) Drop words with a probability of 0.1. (3) Replace 10$\%$ of the input words in $X$ with the [MASK] symbol. (4) Sample a number of token spans from $X$ with span lengths drawn from a Poisson distribution ($\lambda = 3$), and then replace each token span with a single [MASK] token. Here, 0-length spans correspond to the insertion of [MASK] tokens. Based on the performance of different noising strategies (Table 10), we select (4) and use it in pre-training. 
We leave finding better text noising strategies for future work.

We train Unicoder using this task by maximizing the following loss function $\mathcal{L}_{xDAE}$:
\begin{equation}
\mathcal{L}_{xDAE}=\sum_{l_i \in L}\sum_{X \in l_i}\sum_{t=1}^{|X|}\log p(x_t|x_{<t}, c(X)) \nonumber
\end{equation}
where $L={l_1,...,l_N}$ denotes $N$ languages, $X$ is an instance in the $i^{th}$ language $l_i$, $p(x_t|x_{<t}, c(X))$ denotes the probability of generating a single token $x_t$ at time step $t$ given $c(X)$ and $x_{<t}$.

\subsection{Multilingual Future N-gram Prediction (xFNP)}
Motivated by ProphetNet \cite{yan2020prophetnet}, xFNP introduces a future n-gram prediction mechanism to natural language generation. It encourages the model to plan for the future tokens explicitly and prevents over-fitting on strong local correlations.

Given an input text $X = (x_1,x_2,...,x_{|X|}) \in l_i$ from a language $l_i$, 
we randomly mask $k$ token spans of $X$ to generate the masked text $X^{'}$ as the input, and concatenate all masked token spans into $Y$ as the output. Details of this mask strategy are described in Section 6.1. After this,
xFNP first encodes $X^{'}$ to $H_{enc}$ with the encoder:
\begin{equation}
    H_{enc}={\rm Encoder}(X^{'}) \nonumber
\end{equation}
Then, instead of predicting the next token only at each time step, xFNP generates $n$ future tokens simultaneously at time step $t$ with the decoder:
\begin{equation}
\begin{aligned}
    p(y_t|y_{<t},X^{'}),...,p(y_{t+n-1}|y_{<t},X^{'})\\ \nonumber
    ={\rm Decoder}(y_{<t},H_{enc}) \nonumber
\end{aligned}
\end{equation}
%where $1\leq t \leq N-n+1$, $N$ is the number of tokens in the target, $n$ is the number of tokens to be predicted at each time step, and $2 \leq n \leq N$.
Following \citet{yan2020prophetnet}, we set $n=2$.

We train Unicoder using this task by maximizing the following loss function $\mathcal{L}_{xFNP}$:
\begin{equation}
\begin{aligned}
\mathcal{L}_{xFNP}=\sum_{l_i \in L}\sum_{X \in l_i}\{\alpha_0 \cdot \sum_{t=1}^{|Y|}\log p(y_t|y_{<t},X^{'}) \\ \nonumber
+ \alpha_1 \cdot \sum_{t=1}^{|Y|-1}\log p(y_{t+1}|y_{<t},X^{'}) \} \nonumber
\end{aligned}
\end{equation}
where $X^{'}$ and $Y$ are generated from $X$ based on the method mentioned above.
Following \citet{yan2020prophetnet}, we set $\alpha_0=\alpha_1=1$.

\section{Related Work}
\label{sec:rel}
\paragraph{Dataset} GLUE \cite{glue} includes 9 natural language understanding tasks that are labeled in English only. Comparing to GLUE, XGLUE not only expands task annotations to multiple languages, but also includes natural language generation tasks. XNLI \cite{conneau2018xnli}, NER \cite{sang2002ef,sang2003introduction}, POS Tagging \cite{kim-etal-2017-cross}, MLQA \cite{lewis2019mlqa} and PAWS-X \cite{yang2019paws} are 5 multilingual datasets built for specific tasks. XGLUE not only includes these 5 existing tasks, but also introduces 6 new tasks selected from real-world scenarios (i.e., Search, Ads and News). This makes XGLUE have more practical values. XTREME \cite{XTREME} is a concurrent work of XGLUE. Comparing to it, XGLUE includes both understanding and generation tasks, which, to the best of our knowledge, is the first attempt in the cross-lingual dataset construction efforts.

\paragraph{Cross-lingual Pre-trained Model} Multilingual BERT (M-BERT) \cite{devlin-18} performs pre-training based on the multilingual corpus with the masked language model task. By sharing the model parameters and the vocabulary for all languages, M-BERT can obtain the cross-lingual capability over 102 languages. XLM \cite{conneau2019cross} performs cross-lingual pre-training based on multilingual corpus and bilingual corpus, by introducing the translation language model task into pre-training. Based on XLM, Unicoder \cite{huang2019unicoder} uses more cross-lingual pre-training tasks and achieves better results on XNLI. XLM-R \cite{xlmr} is a RoBERTa \cite{liu2019roberta}-version XLM without using translation language model in pre-training. It is trained based on a much larger multilingual corpus (i.e. Common Crawl) and become the new state-of-the-art on XNLI. In this paper, we use both the Common Crawl corpus and the bilingual corpus, aiming to build a stronger baseline model on XGLUE.
BART \cite{lewis2019bart} and ProphetNet \cite{yan2020prophetnet} are two latest generative pre-trained models. We borrow ideas from these two works and extend Unicoder to cross-lingual generation tasks, which goes a step further to verify and explore different text generation approaches in the cross-lingual scenario.

\section{Experiments}
\label{sec:experiments}

\begin{table*}
  \small
	\centering
	\resizebox{1\linewidth}{!}{
	\begin{tabular}{lccccccccccccccccccccc}
		\toprule
		Task & Model & ar & bg & de & el & en & es & fr & hi & it & nl & pl & pt & ru & sw & th & tr & ur & vi & zh & AVG \\
		\midrule
		%\multicolumn{*}{22}{NER}
		\midrule 
		\multirow{3}{*}{NER}                        & M-BERT            & - & - & 69.2 & - & 90.6 & 75.4 & - & - & - & 77.9 & - & - & - & - & - & - & - & - & - & 78.2 \\
		                                            & XLM-R$_{base}$             & - & - & 70.4 & - & 90.9 & 75.2 & - & - & - & 79.5 & - & - & - & - & - & - & - & - & - & 79.0 \\
		                                            & Unicoder$_{LC}$    & - & - & 71.8 & - & 91.1 & 74.4 & - & - & - & 81.6 & - & - & - & - & - & - & - & - & - & \textbf{79.7} \\
		\midrule 
        \multirow{3}{*}{POS}                        & M-BERT    & 52.4& 85.0& 88.7& 81.5& 95.6& 86.8& 87.6& 58.4& 91.3& 88.0& 81.8& 88.3& 78.8& -& 43.3& 69.2& 53.8& 54.3& 58.3& 74.7 \\
		                                            & XLM-R$_{base}$            &67.3& 88.8& 92.2& 88.2& 96.2& 89.0& 89.9& 74.5& 92.6& 88.5& 85.4& 89.7& 86.9& -& 57.9& 72.7& 62.1& 55.2& 60.4& \textbf{79.8}\\
		                                            & Unicoder$_{LC}$  &68.6& 88.5& 92.0& 88.3& 96.1& 89.1& 89.4& 69.9& 92.5& 88.9& 83.6& 89.8& 86.7& -& 57.6& 75.0& 59.8& 56.3& 60.2& 79.6 \\
		\midrule 
		\multirow{3}{*}{NC}                         & M-BERT            & - & - & 82.6 & - & 92.2 & 81.6 & 78.0 & - & - & - & - & - & 79.0 & - & - & - & - & - & - & 82.7 \\
		                                            & XLM-R$_{base}$            & - & - & 84.5 & - & 91.8 & 83.2 & 78.2 & - & - & - & - & - & 79.4 & - & - & - & - & - & - & 83.4 \\
		                                            & Unicoder$_{LC}$    & - & - & 84.2 & - & 91.7 & 83.5 & 78.5 & - & - & - & - & - & 79.7 & - & - & - & - & - & - & \textbf{83.5} \\
		\midrule
        \multirow{3}{*}{MLQA}                       & M-BERT        & 50.9 & - & 63.8 & - & 80.5 & 67.1 & - & 47.9 & - & - & - & - & - & - & - & - & - & 59.5 & 55.4 & 60.7 \\
		                                            & XLM-R$_{base}$         & 56.4 & - & 62.1 & - & 80.1 & 67.9 & - & 60.5 & - & - & - & - & - & - & - & - & - & 67.1 & 61.4 & 65.1 \\
		                                            & Unicoder$_{LC}$& 57.8 & - & 62.7 & - & 80.6 & 68.6 & - & 62.7 & - & - & - & - & - & - & - & - & - & 67.5 & 62.1 & \textbf{66.0} \\
		\midrule 
        \multirow{5}{*}{XNLI}                       & M-BERT            & 64.9                                & 68.9 & 71.1 & 66.4 & 82.1 & 74.3 &                                   73.8 & 60.0 & - & - & - & - & 69.0 &                                   50.4 & 55.8 & 61.6 & 58.0 & 69.5 & 69.3                                & 66.3 \\
		                                            & XLM$\dag$               & 73.1 & 77.4 & 77.8 & 76.6 & 85.0 & 78.9 & 78.7 & 69.6 & - & - & - & - & 75.3 & 68.4 & 73.2 & 72.5 & 67.3 & 76.1 & 76.5 & 75.1 \\
		                                            & XLM-R$_{base}$             & 72.1 & 77.5 & 77.0 & 75.9 & 84.6 & 79.2 & 78.2 & 69.8 & - & - & - & - & 75.5 & 64.7 & 71.6 & 72.9 & 65.1 & 74.8 & 73.7 & 74.2 \\
		                                            & Unicoder$_{SC}$    & 68.5 & 73.2 & 71.6 & 71.6 & 82.9 & 75.0 & 74.7 & 66.0 & - & - & - & - & 70.6 & 64.1 & 67.0 & 68.7 & 62.5 & 71.2 & 69.7 & 70.5 \\
		                                            %& Unicoder$_{LC}$   & 72.7 & 77.8 & 77.5 & 76.4 & 85.6 & 79.3 & 79. 0 & 70.6 & - & - & - & - & 76.3& 66.2& 71.9& 73.1& 65.9& 75.5& 74.2& \textbf{74.8} \\
		                                            & Unicoder$_{LC}$    & 73.9 & 78.5 & 78.2 & 77.3 & 85.4 & 79.8 & 79.2 & 70.1 & - & - & - & - & 76.7 & 67.4 & 71.8 & 73.8 & 66.3 & 75.9 & 74.7 & \textbf{75.3} \\
		                                            %& Unicoder$_{L}^{24}$             & {\color{red} XXX} & {\color{red} XXX} & {\color{red} XXX} & {\color{red} XXX} & {\color{red} XXX} & {\color{red} XXX} & {\color{red} XXX} & {\color{red} XXX} & - & - & - & - & {\color{red} XXX} & {\color{red} XXX} & {\color{red} XXX} & {\color{red} XXX} & {\color{red} XXX} & {\color{red} XXX} & {\color{red} XXX} & {\color{red} XXX} \\
		\midrule 
        \multirow{3}{*}{PAWS-X}                     & M-BERT            & - & - & 82.9 & - & 94.0 & 85.9 & 86.0 & - & - & - & - & - & - & - & - & - & - & - & - & 87.2 \\
		                                            & XLM-R$_{base}$             & - & - & 86.9 & - & 94.4 & 88.0 & 88.7 & - & - & - & - & - & - & - & - & - & - & - & - & 89.5 \\
		                                            & Unicoder$_{LC}$ & - & - & 87.4 & - & 94.9 & 88.8 & 89.3 & - & - & - & - & - & - & - & - & - & - & - & - & \textbf{90.1}\\
		\midrule 
        \multirow{3}{*}{QADSM}                      & M-BERT            & - & - & 60.3 & - & 68.3 & - & 64.1 & - & - & - & - & - & - & - & - & - & - & - & - & 64.2 \\
		                                            & XLM-R$_{base}$             & - & - & 65.8 & - & 71.7 & - & 68.3 & - & - & - & - & - & - & - & - & - & - & - & - & \textbf{68.6} \\
		                                            & Unicoder$_{LC}$   &- & - & 64.6 & - & 71.8 & - & 68.7 & - & - & - & - & - & - & - & - & - & - & - & - & 68.4\\
        \midrule 
        \multirow{3}{*}{WPR}                        & M-BERT            &- & - & 76.6 & - & 78.1 & 75.3 & 74.2 & - & 70.1 & - & - & 76.6 & - & - & - & - & - & - & 64.5 & 73.5 \\
		                                            & XLM-R$_{base}$           & - & - & 77.6 & - & 78.2 & 76.0 & 74.4 & - & 70.7 & - & - & 77.3 & - & - & - & - & - & - & 63.9 & 73.8 \\
		                                            & Unicoder$_{LC}$   & - & - & 77.2 & - & 78.4 & 75.7 & 74.9 & - & 70.3 & - & - & 77.4 & - & - & - & - & - & - & 64.4 & \textbf{73.9} \\
		\midrule 
        \multirow{3}{*}{QAM}                        
                                                    & M-BERT            & - & - & 64.7 & - & 67.5 & - & 66.0 & - & - & - & - & - & - & - & - & - & - & - & - & 66.1 \\
		                                            & XLM-R$_{base}$              & - & - & 68.1 & - & 69.3 & - & 67.8 & - & - & - & - & - & - & - & - & - & - & - & - & 68.4 \\
		                                            & Unicoder$_{LC}$   &  - & - & 68.4 & - & 69.9 & - & 68.4 & - & - & - & - & - & - & - & - & - & - & - & - & \textbf{68.9}\\
        \bottomrule
		\multirow{3}{*}{AVG$_U^2$}                  & M-BERT &  &  &  &  &  &  &  &  &  &  &  &  &  &  &  &  & & &  & 72.6 \\
		                                            & XLM-R$_{base}$ &  &  &  &  &  &  &  &  &  &  &  &  &  &  &  &  & & &  & 75.8 \\
		                                            & Unicoder$_{LC}$ &  &  &  &  &  &  &  &  &  &  &  &  &  & & &  &  &  &  & \textbf{76.2} \\
        \bottomrule
		\bottomrule
        \multirow{4}{*}{QG}                         & M-BERT            & - & - & 0.1 & - & 7.8 & 0.1 & 0.1 & - & 0.2 & - & - & 0.1 & - & - & - & - & - & - & - & 1.4 \\
		                                            & XLM-R$_{base}$               & - & - & 0.1 & - & 6.0 & 0.0 & 0.0 & - & 0.1 & - & - & 0.0 & - & - & - & - & - & - & - & 1.0 \\
		                                            & Unicoder$_{SC}^{xDAE}$    & - & - & 3.0 & - & 14.0 & 12.4 & 4.2 & - & 15.8 & - & - & 8.3 & - & - & - & - & - & - & - & 9.6 \\
		                                            & Unicoder$_{SC}^{xFNP}$    & - & - & 3.7 & - & 13.9 & 14.8 & 4.9 & - & 17.0 & - & - & 9.5 & - & - & - & - & - & - & - & \textbf{10.6} \\
		\midrule 
        \multirow{4}{*}{NTG}                        & M-BERT            & - & - & 0.7 & - & 9.0 & 0.4 & 0.4 & - & - & - & - & - & 0.0 & - & - & - & - & - & - & 2.1 \\
		                                            & XLM-R$_{base}$               & - & - & 0.6 & - & 8.1 & 0.4 & 0.3 & - & - & - & - & - & 0.0 & - & - & - & - & - & - & 1.9 \\
		                                            & Unicoder$_{SC}^{xDAE}$    & - & - & 6.8 & - & 15.6 & 9.0 & 8.7 & - & - & - & - & - & 7.7 & - & - & - & - & - & - & 9.6 \\
		                                            & Unicoder$_{SC}^{xFNP}$    & - & - & 7.5 & - & 15.8 & 11.9 & 9.9 & - & - & - & - & - & 8.4 & - & - & - & - & - & - & \textbf{10.7} \\
		
		\bottomrule
		\multirow{4}{*}{AVG$_G^2$}                  & M-BERT &  &  &  &  &  &  &  &  &  &  &  &  &  &  &  &  & & &  & 1.8 \\
		                                            & XLM-R$_{base}$ &  &  &  &  &  &  &  &  &  &  &  &  &  &  &  &  & & &  & 1.5 \\
		                                            & Unicoder$_{SC}^{xDAE}$ &  &  &  &  &  &  &  &  &  &  &  &  &  & & &  &  &  &  & 9.6 \\
		                                            & Unicoder$_{SC}^{xFNP}$ &  &  &  &  &  &  &  &  &  &  &  &  &  & & &  &  &  &  & \textbf{10.7} \\
		
		\bottomrule
		\bottomrule
	\end{tabular}
	}
	\caption{The overall evaluation results on XGLUE. We use M-BERT \cite{devlin-18}, XLM \cite{conneau2019cross} and XLM-R$_{base}$ \cite{xlmr} as baselines. Unicoder$_{SC}$ and Unicoder$_{LC}$ are pre-trained using small corpus and large corpus, respectively. 
	Unicoder$_{SC}^{xDAE}$ and Unicoder$_{SC}^{xFNP}$ are pre-trained by xDAE and xFNP for 100 languages, respectively. 
	For the results of M-BERT/XLM-R on generation tasks, we initialize the encoder-decoder model with M-BERT/XLM-R and fine-tune it on each downstream task without pre-training.  
	\textbf{All models are (12-layer) base ones.}
	Given a task, each pre-trained model is fine-tuned using its English training set only, and then applied to all test sets in different languages. AVG$_U^2$ and AVG$_G^2$ denote the average score of the average scores on 9 understanding tasks and 2 generation tasks, respectively.
	Due to GPU limitation and time cost consideration, 
	Unicoder$_{LC}$ is pre-trained using 10$\%$ of the large corpus only.
	}
	\label{tab:main-results} 
\end{table*}

\begin{table*}
\small
    \begin{center}
          \resizebox{1\linewidth}{!}{
    	\begin{tabular}{ccccccccccccccccc}
        	\toprule
          Pivot	& en & fr & es & de & el & bg & ru & tr & ar & vi & th & zh & hi & sw & ur & AVG  \\
            \midrule
            \midrule
en & \textbf{85.4} & 79.2 & 79.8 & 78.2 & 77.3 & 78.5 & 76.7 & 73.8 & 73.9 & 75.9 & 71.8 & 74.7 & 70.1 & 67.4 & 66.3 & 75.3\\
%en & \textbf{85.6} & 79.2 & 79.4 & 77.4 & 76.8 & 78.3 & 76.4 & 72.8 & 73.6 & 75.7 & 71.5 & 74.5 & 70.4 & 66.3 & 66.0 & 74.9\\
fr & 84.0 & 79.9 & 80.3 & 78.8 & 77.4 & 79.2 & 77.0 & 73.6 & 73.7 & 76.7 & 72.7 & 75.3 & 73.0 & 67.4 & 68.3 & 75.8\\
es & 84.5 & \textbf{80.2} & \textbf{81.2} & 79.7 & 78.2 & 79.2 & 77.6 & 74.5 & 74.8 & 77.0 & 72.8 & 76.2 & 73.2 & 67.7 & \textbf{69.6} & \textbf{76.4}\\
de & 83.5 & 79.1 & 80.1 & \textbf{80.2} & 77.9 & 78.6 & 77.0 & 74.9 & 74.6 & 76.1 & 73.3 & 76.2 & 73.1 & 67.7 & 68.9 & 76.1\\
el & 83.8 & 80.1 & 81.0 & 78.6 & \textbf{79.6} & 79.3 & 77.0 & 74.2 & 74.9 & 77.1 & 73.5 & 75.9 & 72.7 & 69.1 & 69.1 & \textbf{76.4}\\
bg & 83.5 & 79.6 & 80.4 & 79.1 & 77.9 & \textbf{80.5} & 77.9 & 74.9 & 73.9 & 76.5 & 73.9 & 75.6 & 72.8 & 68.6 & 68.9 & 76.3\\
ru & 84.1 & 79.9 & 79.9 & 78.8 & 77.5 & 79.9 & \textbf{78.1} & 73.9 & 74.5 & 77.1 & 73.8 & 75.7 & 73.1 & 68.5 & 69.0 & 76.2\\
tr & 83.3 & 78.4 & 79.6 & 78.4 & 77.5 & 79.2 & 77.5 & \textbf{77.1} & 74.2 & 77.1 & 74.5 & 76.5 & 73.7 & 69.3 & 70.3 & \textbf{76.4}\\
ar & 83.2 & 78.9 & 79.5 & 77.6 & 77.4 & 78.6 & 77.0 & 75.4 & \textbf{76.8} & 76.8 & 74.0 & 76.0 & 73.0 & 69.5 & 69.3 & 76.2\\
vi & 83.2 & 78.6 & 79.1 & 77.7 & 76.6 & 78.9 & 77.5 & 75.3 & 74.7 & \textbf{78.5} & 73.5 & \textbf{76.8} & 73.1 & 67.8 & 69.0 & 76.0\\
th & 82.5 & 78.5 & 79.1 & 77.8 & 77.1 & 78.3 & 76.7 & 75.0 & 74.3 & 76.9 & \textbf{76.4} & 76.2 & 72.9 & 68.4 & 69.7 & 76.0\\
zh & 81.6 & 78.2 & 77.9 & 77.1 & 76.0 & 77.9 & 76.2 & 73.7 & 73.7 & 75.8 & 73.6 & 76.6 & 71.7 & 67.4 & 68.3 & 75.1\\
hi & 81.8 & 78.5 & 79.2 & 76.7 & 77.2 & 78.2 & 76.2 & 74.5 & 73.9 & 76.4 & 71.7 & 75.2 & \textbf{73.8} & 68.2 & 68.5 & 75.3\\
sw & 82.0 & 77.6 & 78.8 & 77.2 & 76.5 & 77.7 & 76.2 & 74.4 & 74.3 & 76.3 & 74.0 & 75.2 & 72.2 & \textbf{71.4} & 69.5 & 75.6\\
ur & 76.7 & 72.5 & 74.1 & 72.6 & 72.1 & 73.9 & 72.7 & 69.7 & 69.7 & 72.8 & 70.1 & 72.4 & 69.0 & 66.0 & 67.5 & 71.5\\
        	\bottomrule
        \end{tabular}
        }
        \end{center}
    \caption{Impacts of different pivot languages on XNLI. Given each pivot language, the corresponding fine-tuned XNLI results on all languages are listed in the same row. Each \textbf{bolded number} is the best result in that column.}
\end{table*}

\subsection{Experimental Settings}
\paragraph{Understanding Tasks}
The hyper-parameters are set as follows: 768 hidden units, 12 heads, GELU activation, a dropout rate of 0.1, 512 max input length, 12 layers in encoder.

In the pre-training stage, we first initialize Unicoder$_{LC}$ with XLM-R$_{base}$ \cite{xlmr}, and then run continue pre-training with the accumulated 8,192 batch size with gradients accumulation. We use Adam Optimizer with a linear warm-up \cite{vaswani2017attention} and set the learning rate to 3e-5. We select different understanding tasks randomly in different batches.

In the fine-tuning stage, the batch size is set to 32. We use Adam Optimizer~\cite{kingma2014adam} with warm-up and set the learning rate to 5e-6. For all sentence classification tasks, we fine-tune 10 epochs. For POS Tagging and NER, we fine-tune 20 epochs. And for POS Tagging, we set the learning rate to 2e-5. For MLQA, we set the learning rate to 3e-5, batch size to 12 and train 2 epochs following BERT for SQuAD.
After each epoch, we test the fine-tuned model on the dev sets of all languages. 
We select the model with the best average result on the dev sets of all languages.

\paragraph{Generation Tasks}
We evaluate Unicoder$_{SC}^{xDAE}$ and Unicoder$_{SC}^{xFNP}$ as two separate models.

For Unicoder$_{SC}^{xDAE}$, the hyper-parameters are set as follows: 768 hidden units, 12 heads, GELU activation, a dropout rate of 0.1, 512 max input length, 12 layers in encoder, 12 layers in decoder.

In the pre-training stage, we first initialize encoder and decoder with XLM-R \cite{xlmr}, and then run continue pre-training with 1,024 batch size. We use Adam optimizer with warm-up and set the learning rate to 2e-4.

In the fine-tuning stage, the batch size is 1024. We use Adam Optimizer \cite{kingma2014adam} with learning rate 1e-5 and warm-up steps 2000.

For Unicoder$_{SC}^{xFNP}$, the hyper-parameters are set as follows:
1,024 hidden size, 12 layers in encoder, 12 layers in decoder, 512 max input length.

In the pre-training stage, we pre-train the model from scratch and follow ProphetNet \cite{yan2020prophetnet} to randomly mask a continuous span (with a fixed length 9) in every 64 tokens. About 15\% of the tokens in original sequence are masked in this step. We use a special symbol [MASK] to replace 80\% of the masked tokens, keep 10\% unchanged, and random replace 10\% of the masked tokens. We set the batch size to 1,024, training steps to 350,000. The learning rate is set to 1e-4. We set the number of future tokens $n$ to 2.

In the fine-tuning stage, we use Adam Optimizer \cite{kingma2014adam} and set the learning rate to 1e-4. We set the batch size to 64 and the warm-up steps to 1,000.

\begin{table}
  \small
	\centering
	\resizebox{1\linewidth}{!}{
	\begin{tabular}{ccccccccc}
		\toprule
		Pivot & en & es & fr & de & ru & AVG \\
		\midrule
		\midrule
		en & \textbf{15.6}/\textbf{15.8} & 9.0/11.9 & 8.7/9.9 & 6.8/7.5 & 7.7/8.4 & 9.6/10.7 \\
        es & 7.8/8.8 & \textbf{17.1}/\textbf{17.1} & 10.6/10.9 & 7.6/8.0 & 8.0/8.6 & 10.2/10.7 \\
        fr & 8.2/8.7 & 11.4/12.5 & \textbf{19.4}/\textbf{20.9} & 8.3/8.2 & 7.6/7.8 & \textbf{11.0}/\textbf{11.6} \\
        de & 8.2/8.6 & 9.9/11.2 & 9.5/10.2 &\textbf{14.1}/\textbf{13.7} & 8.4/8.0 & 10.0/10.3 \\
        ru & 6.9/7.4 & 9.3/10.8 & 8.8/9.9 & 6.9/7.0 & \textbf{16.6}/\textbf{16.7} & 9.7/10.4 \\
        \bottomrule
	\end{tabular}
	}
	\caption{Impacts of different pivot languages on NTG. 
	Unicoder$_{SC}^{xDAE}$/Unicoder$_{SC}^{xFNP}$ evaluated by BLEU-4.}
	\label{tab:newszero} 
\end{table}

\begin{table*} [t]
    \small
    \begin{center}
            \resizebox{1\linewidth}{!}{
    	\begin{tabular}{lcccccccccccccccc}
        	\toprule
        	& en & fr & es & de & el & bg & ru & tr & ar & vi & th & zh & hi & sw & ur & AVG \\
            \midrule
            \midrule
            XLM-R$_{base}\ (pl)$ &  84.6 & 78.2 & 79.2 & 77.0 & 75.9 & 77.5 & 75.5 & 72.9 & 72.1 & 74.8 & 71.6 & 73.7 & 69.8 & 64.7 & 65.1 & 74.2 \\
            XLM-R$_{base}\ (ml)$ & 85.7 & 81.5 & \textbf{82.5} & 81.2 & 79.7 & 81.7 & \textbf{80.0} & \textbf{79.0} & 77.1 & 80.1 & 77.9 & 79.2 & \textbf{76.5} & 73.0 & 71.3 & 79.1 \\
            \midrule
            Unicoder$_{LC}\ (pl)$ & 85.4 & 79.2 & 79.8 & 78.2 & 77.3 & 78.5 & 76.7 & 73.8 & 73.9 & 75.9 & 71.8 & 74.7 & 70.1 & 67.4 & 66.3 & 75.3 \\
        	Unicoder$_{LC}\ (ml)$ & \textbf{85.8} & \textbf{81.9} & 82.3 & \textbf{81.5} & \textbf{80.8} & \textbf{82.0} & 79.9 & 78.7 & \textbf{78.1} & \textbf{80.2} & \textbf{78.4} & \textbf{79.3} & 76.2 & \textbf{73.2} & \textbf{72.4} & \textbf{79.4} \\
        	\bottomrule
        \end{tabular}
        }
        \end{center}
    \caption{Impact of multi-language fine-tuning on XNLI. $pl$ and $ml$ denote pivot-language fine-tuning (English as pivot) and multi-language fine-tuning, respectively. }
\end{table*}

\begin{table} [t]
  \small
	\centering
	\resizebox{1\linewidth}{!}{
	\begin{tabular}{lcccccccc}
		\toprule
		Model & en & es & fr & de & ru & AVG \\
		\midrule
		\midrule
		Unicoder$_{SC}^{xDAE}\ (pl)$ & 15.6 & 9.0 & 8.7 & 6.8 & 7.7 & 9.6 \\
        Unicoder$_{SC}^{xDAE}\ (ml)$ & \textbf{18.5}  & \textbf{18.3} & \textbf{28.2} & \textbf{15.5} & \textbf{33.4} & \textbf{22.8} \\
        \midrule
        Unicoder$_{SC}^{xFNP}\ (pl)$ & \textbf{15.8} & 11.9 & 9.9 & 7.5 & 8.4 & 10.7 \\
        Unicoder$_{SC}^{xFNP}\ (ml)$ & 15.6 & \textbf{17.1} & \textbf{19.1} & \textbf{13.9} & \textbf{15.8} & \textbf{16.3} \\
        \bottomrule
	\end{tabular}
	}
	\caption{Impact of multi-language fine-tuning on NTG. $pl$ and $ml$ denote pivot-language fine-tuning (English as pivot) and multi-language fine-tuning, respectively. BLUE-4 is the metric.}
	\label{tab:newsall} 
\end{table}

\begin{table} [t]
  \small
	\centering
	\resizebox{1\linewidth}{!}{
	\begin{tabular}{lccccccc}
		\toprule
		Model & XNLI & PAWS-X & NC & QAM & QADSM & AVG \\
		\midrule
		\midrule
		Unicoder$_{LC}\ (pl)$ & \textbf{75.3} & 90.1 & \textbf{83.5} & \textbf{68.9} & 68.4 & \textbf{77.2} \\
        Unicoder$_{LC}\ (mt)$ & 74.4 & \textbf{90.2} &83.4 & 68.7 & \textbf{69.0} & 77.1 \\
        \bottomrule
    	\end{tabular}
	}
	\caption{Impacts of multi-task fine-tuning on XNLI, PAWS-X, NC, QAM and QADSM. $pl$ and $mt$ denote pivot-language fine-tuning (English as pivot) on each task and multi-task fine-tuning, respectively.}
	\label{tab:newszero} 
\end{table}

\begin{table} [t]
  \small
	\centering
	\resizebox{1\linewidth}{!}{
	\begin{tabular}{lcccccccc}
		\toprule
		Text Noising Strategy & en & es & fr & de & ru & AVG \\
		\midrule
		\midrule
		(1)+(2)+(3) & 14.6 & 8.5 & 7.4 & 6.0 & 7.4 & 8.8 \\
		(4) & 14.8 & \textbf{8.7} & \textbf{7.5} & \textbf{6.7} & \textbf{8.2} & \textbf{9.2} \\
		(1)+(2)+(3)+(4) & \textbf{15.2} & 7.9  & 7.3& 6.2 & 7.7 & 8.9 \\
        \bottomrule
	\end{tabular}
	}
	\caption{Impact of different text noising strategies on NTG using pivot-language fine-tuning (English as pivot). BLUE-4 is the metric.}
	\label{tab:newsall} 
\end{table}

\begin{table} [t]
  \small
	\centering
	%\resizebox{1\linewidth}{!}{
	\begin{tabular}{lccc}
		\toprule
		Model & fr & zh & AVG \\
		\midrule
		\midrule
		XNLG \cite{chi2019cross}   & 36.3 & 38.9 & 37.6 \\
        Unicoder$_{SC}^{xDAE}$     & \textbf{37.9} & \textbf{42.2} & \textbf{40.1} \\
        \bottomrule
	\end{tabular}
	%}
	\caption{The zero-shot results on Abstractive Summarization. Unicoder$_{SC}^{xDAE}$ and XNLG are fine-tuned using English labeled data. ROUGE-L is the metric.}
	\label{tab:newszero} 
\end{table}

\subsection{Main Result}

7 cross-lingual pre-trained models are evaluated on XGLUE and compared in Table 4: 
12-layer M-BERT \cite{devlin-18} trained on Wikipedia corpus for 102 languages, 
12-layer XLM \cite{conneau2019cross} trained on Wikipedia and bilingual corpora for 15 languages, 
12-layer XLM-R$_{base}$ \cite{xlmr} trained on Common Crawl corpus for 100 languages, 
12-layer Unicoder$_{SC}$ trained on small corpus for 100 languages, 
12-layer Unicoder$_{LC}$ trained on large corpus for 100 languages,
12-layer Unicoder$_{SC}^{xDAE}$ and 12-layer Unicoder$_{SC}^{xFNP}$ trained on Wikipedia corpus for 100 languages.
Given a downstream task, each pre-trained model is fine-tuned using its English training set and then applied to all test sets in different languages.
Note that, all results are reproduced by this paper, except the XLM$\dag$ results on XNLI are from \citet{conneau2019cross}.
%As XLM-R$_{base}$ (or Unicoder$_{LC}$) performs consistently better than XLM (or Unicoder$_{SC}$), we discard the results of XLM and Unicoder$_{SC}$ on all tasks except XNLI. 

We find 
(1) Unicoder$_{LC}$ performs slightly better than M-BERT and XLM-R$_{base}$ on the 9 understanding tasks, as it is pre-trained based on multilingual and bilingual corpora at the same time and uses TLM;
(2) Unicoder$_{LC}$ performs better than Unicoder$_{SC}$, as it is pre-trained based on the larger corpus;
(3) Unicoder$_{SC}^{xDAE}$ and Unicoder$_{SC}^{xFNP}$ show good cross-lingual transfer capabilities and perform significantly better than M-BERT and XLM-R$_{base}$ on the 2 generation tasks. 
It proves the importance of introducing generation tasks into pre-training for cross-lingual text generation;
(4) Unicoder$_{SC}^{xFNP}$ performs slightly better than Unicoder$_{SC}^{xDAE}$. 
But it is not a fair comparison, because they use different text denoising tasks (sentence prediction vs. span prediction) and different generation mechanisms (single-token prediction vs. multi-token prediction). 
We leave combining these two tasks for future work.

\subsection{Ablation Study}

\subsubsection{Pivot-language Fine-tuning}
We define pivot-language (\textit{pl}) fine-tuning as fine-tune a pre-trained model for a downstream task using its labeled data in a pivot language (e.g. English) and the apply the fine-tuned model to all languages. Table 4 chooses English as the pivot language, as all tasks in XGLUE have labeled data in English. But is English always the optimal choice? Will the results become better, if we do fine-tuning using other pivot languages?

To answer these questions, we evaluate Unicoder on XNLI and NTG using different pivot languages in fine-tuning and list comparison results in Table 5 and Table 6, respectively. 
(1) For each test set in language $l_i$ in Table 5 and Table 6, its best result is often achieved when the model is fine-tuned using $l_i$ as the pivot language; 
(2) For XNLI in Table 5, the best pivot languages are Spanish (es), Greek (el) and Turkish (tr), rather than English (en). For NTG in Table 6, the best pivot language is French (fr) for both Unicoder$_{SC}^{xDAE}$ and Unicoder$_{SC}^{xFNP}$. 
It means the average quality of a cross-lingual pre-trained model could be further improved on a  downstream task, by selecting a specific pivot language in fine-tuning.
%\textit{We leave explorations on pivot languages for future work.}

\subsubsection{Multi-language Fine-tuning}
We define multi-language (\textit{ml}) fine-tuning as fine-tune a pre-trained model for a downstream task using all its available labeled data in different languages. We evaluate Unicoder on XNLI and NTG using this fine-tuning method and list evaluation results in Table 7 and Table 8, respectively.

We find multi-language fine-tuning can achieve better results than pivot-language fine-tuning on both XNLI and NTG. 
It means the average quality of a cross-lingual pre-trained model could be significantly improved on a downstream task, by using combined labeled data in multiple languages.

\subsubsection{Multi-task Fine-tuning}

We define multi-task fine-tuning as fine-tune a pre-trained model for multiple downstream tasks using their combined labeled data.
To reduce the experimental cost, we evaluate Unicoder on following 5 understanding tasks: XNLI, PAWS-X, NC, QAM and QADSM, using their merged English labeled data in fine-tuning.
Results are listed in Table 9.

We find PAWS-X and QADSM can benefit from the joint fine-tuning strategy, but XNLI, NC and QAM cannot. 
We leave discovering relationships between different tasks for better downstream task fine-tuning for future work.

\subsubsection{Impacts of Text Noising Strategies}
We investigate the impacts of different text noising strategies (Section 4.1) in Unicoder$_{SC}^{xDAE}$, and list comparison results in Table 10, where (1)+(2)+(3) denotes the result of using the first three strategies in pre-training, (4) denotes the result of using the last strategy in pre-training, (1)+(2)+(3)+(4) denotes the result of using all strategies in pre-training. 
To reduce experiment cost, we set max sequence length to 256 and only train 60K steps.
We find that (4) can achieve the best average result on NTG. So all results of Unicoder$_{SC}^{xDAE}$ reported in this paper is pre-trained using (4) only.
%Table 10 gives some input-output examples of Unicoder$_{SC}^{xDAE}$ on NTG.

We also compare Unicoder$_{SC}^{xDAE}$ with XNLG \cite{chi2019cross} on the Abstractive Summarization task. For fairly comparison, we implement xDAE in same code base and use same pre-training languages as XNLG. 
The zero-shot comparison results are listed in Table 11. We can see that by using xDAE only in pre-training, Unicoder$_{SC}^{xDAE}$ can outperform XNLG significantly, which is pre-trained using 4 tasks including MLM, DAE, XMLM and XAE. 
It verifies the effectiveness of the fourth text noising strategy described in Section 4.1 for generation tasks.

%\subsection{Updates in the Next Version}
%We will add 3 updates in the next version: 
%(1) the results of a 24-layer Unicoder on understanding tasks; 
%(2) the results of a 12-layer Unicoder on understanding tasks, which is pre-trained by new tasks beyond MLM and TLM; 
%(3) the comparison results of Unicoder$_{SC}^{xDAE}$ and Unicoder$_{SC}^{xFNP}$ on generation tasks, which are pre-trained based on the small corpus for 100 languages.
\section{Conclusion}
\label{sec:conclusion}
We present XGLUE as a new cross-lingual benchmark and conduct comprehensive evaluations with interesting findings observed. 
%We hope it can advance the developments of cross-lingual research and applications.
We thank STC-A NLP, Bing Answers, Bing Ads, Bing Relevance and Microsoft News for providing the datasets.

\bibliography{bib/journal-abbrv,bib/ref}

\begin{thebibliography}{29}
\expandafter\ifx\csname natexlab\endcsname\relax\def\natexlab#1{#1}\fi

\bibitem[{Chen et~al.(2019)Chen, Li, Yu, Kholy, Ahmed, Gan, Cheng, and
  Liu}]{uniter}
Yen-Chun Chen, Linjie Li, Licheng Yu, Ahmed~El Kholy, Faisal Ahmed, Zhe Gan,
  Yu~Cheng, and Jingjing Liu. 2019.
\newblock Uniter: Learning universal image-text representations.
\newblock \emph{arXiv}.

\bibitem[{Chi et~al.(2019)Chi, Dong, Wei, Wang, Mao, and Huang}]{chi2019cross}
Zewen Chi, Li~Dong, Furu Wei, Wenhui Wang, Xian-Ling Mao, and Heyan Huang.
  2019.
\newblock Cross-lingual natural language generation via pre-training.
\newblock In \emph{AAAI}.

\bibitem[{Conneau et~al.(2019)Conneau, Khandelwal, Goyal, Chaudhary, Wenzek,
  Guzm{\'a}n, Grave, Ott, Zettlemoyer, and Stoyanov}]{xlmr}
Alexis Conneau, Kartikay Khandelwal, Naman Goyal, Vishrav Chaudhary, Guillaume
  Wenzek, Francisco Guzm{\'a}n, Edouard Grave, Myle Ott, Luke Zettlemoyer, and
  Veselin Stoyanov. 2019.
\newblock Unsupervised cross-lingual representation learning at scale.
\newblock \emph{arXiv}.

\bibitem[{Conneau and Lample(2019)}]{conneau2019cross}
Alexis Conneau and Guillaume Lample. 2019.
\newblock Cross-lingual language model pretraining.
\newblock In \emph{NeurIPS}.

\bibitem[{Conneau et~al.(2018)Conneau, Lample, Rinott, Williams, Bowman,
  Schwenk, and Stoyanov}]{conneau2018xnli}
Alexis Conneau, Guillaume Lample, Ruty Rinott, Adina Williams, Samuel~R Bowman,
  Holger Schwenk, and Veselin Stoyanov. 2018.
\newblock Xnli: Evaluating cross-lingual sentence representations.
\newblock \emph{arXiv preprint arXiv:1809.05053}.

\bibitem[{Devlin et~al.(2019)Devlin, Chang, Lee, and Toutanova}]{devlin-18}
Jacob Devlin, Ming-Wei Chang, Kenton Lee, and Kristina Toutanova. 2019.
\newblock {BERT}: Pre-training of deep bidirectional transformers for language
  understanding.
\newblock In \emph{NAACL}.

\bibitem[{Dong et~al.(2019)Dong, Yang, Wang, Wei, Liu, Wang, Gao, Zhou, and
  Hon}]{unilm}
Li~Dong, Nan Yang, Wenhui Wang, Furu Wei, Xiaodong Liu, Yu~Wang, Jianfeng Gao,
  Ming Zhou, and Hsiao-Wuen Hon. 2019.
\newblock Unified language model pre-training for natural language
  understanding and generation.
\newblock \emph{NeurIPS}.

\bibitem[{Hu et~al.(2020)Hu, Ruder, Siddhant, Neubig, Firat, and
  Johnson}]{XTREME}
Junjie Hu, Sebastian Ruder, Aditya Siddhant, Graham Neubig, Orhan Firat, and
  Melvin Johnson. 2020.
\newblock Xtreme: A massively multilingual multi-task benchmark for evaluating
  cross-lingual generalization.
\newblock \emph{arXiv}.

\bibitem[{Huang et~al.(2019)Huang, Liang, Duan, Gong, Shou, Jiang, and
  Zhou}]{huang2019unicoder}
Haoyang Huang, Yaobo Liang, Nan Duan, Ming Gong, Linjun Shou, Daxin Jiang, and
  Ming Zhou. 2019.
\newblock Unicoder: A universal language encoder by pre-training with multiple
  cross-lingual tasks.
\newblock In \emph{EMNLP}.

\bibitem[{Kim et~al.(2017)Kim, Kim, Sarikaya, and
  Fosler-Lussier}]{kim-etal-2017-cross}
Joo-Kyung Kim, Young-Bum Kim, Ruhi Sarikaya, and Eric Fosler-Lussier. 2017.
\newblock Cross-lingual transfer learning for {POS} tagging without
  cross-lingual resources.
\newblock In \emph{Proceedings of the 2017 Conference on Empirical Methods in
  Natural Language Processing}, pages 2832--2838, Copenhagen, Denmark.
  Association for Computational Linguistics.

\bibitem[{Kingma and Ba(2014)}]{kingma2014adam}
Diederik~P Kingma and Jimmy Ba. 2014.
\newblock Adam: A method for stochastic optimization.
\newblock \emph{arXiv preprint arXiv:1412.6980}.

\bibitem[{Lewis et~al.(2019{\natexlab{a}})Lewis, Liu, Goyal, Ghazvininejad,
  Mohamed, Levy, Stoyanov, and Zettlemoyer}]{lewis2019bart}
Mike Lewis, Yinhan Liu, Naman Goyal, Marjan Ghazvininejad, Abdelrahman Mohamed,
  Omer Levy, Ves Stoyanov, and Luke Zettlemoyer. 2019{\natexlab{a}}.
\newblock Bart: Denoising sequence-to-sequence pre-training for natural
  language generation, translation, and comprehension.
\newblock \emph{arXiv preprint arXiv:1910.13461}.

\bibitem[{Lewis et~al.(2019{\natexlab{b}})Lewis, O{\u{g}}uz, Rinott, Riedel,
  and Schwenk}]{lewis2019mlqa}
Patrick Lewis, Barlas O{\u{g}}uz, Ruty Rinott, Sebastian Riedel, and Holger
  Schwenk. 2019{\natexlab{b}}.
\newblock Mlqa: Evaluating cross-lingual extractive question answering.
\newblock \emph{arXiv preprint arXiv:1910.07475}.

\bibitem[{Li et~al.(2020)Li, Duan, Fang, Gong, Jiang, and Zhou}]{unicodervl}
Gen Li, Nan Duan, Yuejian Fang, Ming Gong, Daxin Jiang, and Zhou Zhou. 2020.
\newblock Unicoder-vl: A universal encoder for vision and language by
  cross-modal pre-training.
\newblock \emph{AAAI}.

\bibitem[{Liu et~al.(2019)Liu, Ott, Goyal, Du, Joshi, Chen, Levy, Lewis,
  Zettlemoyer, and Stoyanov}]{liu2019roberta}
Yinhan Liu, Myle Ott, Naman Goyal, Jingfei Du, Mandar Joshi, Danqi Chen, Omer
  Levy, Mike Lewis, Luke Zettlemoyer, and Veselin Stoyanov. 2019.
\newblock Roberta: A robustly optimized bert pretraining approach.
\newblock \emph{arXiv preprint arXiv:1907.11692}.

\bibitem[{Lu et~al.(2019)Lu, Batra, Parikh, and Lee}]{vilbert}
Jiasen Lu, Dhruv Batra, Devi Parikh, and Stefan Lee. 2019.
\newblock Vilbert: Pretraining task-agnostic visiolinguistic representations
  for vision-and-language tasks.
\newblock \emph{NeurIPS}.

\bibitem[{Radford et~al.(2018)Radford, Narasimhan, Salimans, and
  Sutskever}]{gpt}
Alec Radford, Karthik Narasimhan, Tim Salimans, and Ilya Sutskever. 2018.
\newblock Improving language understanding by generative pre-training.
\newblock \emph{arXiv}.

\bibitem[{Sang and De~Meulder(2003)}]{sang2003introduction}
Erik Tjong~Kim Sang and Fien De~Meulder. 2003.
\newblock Introduction to the conll-2003 shared task: Language-independent
  named entity recognition.
\newblock In \emph{Proceedings of the Seventh Conference on Natural Language
  Learning at HLT-NAACL 2003}, pages 142--147.

\bibitem[{Sang(2002)}]{sang2002ef}
Tjong~Kim Sang. 2002.
\newblock Ef: Introduction to the conll-2002 shared task.
\newblock In \emph{Proceedings of the 6th Conference on Natural Language
  Learning}.

\bibitem[{Schwenk et~al.(2019)Schwenk, Wenzek, Edunov, Grave, and
  Joulin}]{ccmatrix}
Holger Schwenk, Guillaume Wenzek, Sergey Edunov, Edouard Grave, and Armand
  Joulin. 2019.
\newblock Ccmatrix: Mining billions of high-quality parallel sentences on the
  web.
\newblock \emph{arXiv}.

\bibitem[{Vaswani et~al.(2017)Vaswani, Shazeer, Parmar, Uszkoreit, Jones,
  Gomez, Kaiser, and Polosukhin}]{vaswani2017attention}
Ashish Vaswani, Noam Shazeer, Niki Parmar, Jakob Uszkoreit, Llion Jones,
  Aidan~N Gomez, {\L}ukasz Kaiser, and Illia Polosukhin. 2017.
\newblock Attention is all you need.
\newblock In \emph{NeurIPS}.

\bibitem[{Wang et~al.(2019)Wang, Singh, Michael, Hill, Levy, and Bowman}]{glue}
Alex Wang, Amanpreet Singh, Julian Michael, Felix Hill, Omer Levy, and Samuel
  Bowman. 2019.
\newblock Glue: A multi-task benchmark and analysis platform for natural
  language understanding.
\newblock \emph{ICLR}.

\bibitem[{Wenzek et~al.(2019)Wenzek, Lachaux, Conneau, Chaudhary, Guzman,
  Joulin, and Grave}]{wenzek2019ccnet}
Guillaume Wenzek, Marie-Anne Lachaux, Alexis Conneau, Vishrav Chaudhary,
  Francisco Guzman, Armand Joulin, and Edouard Grave. 2019.
\newblock Ccnet: Extracting high quality monolingual datasets from web crawl
  data.
\newblock \emph{arXiv preprint arXiv:1911.00359}.

\bibitem[{Yan et~al.(2020)Yan, Qi, Gong, Liu, Duan, Chen, Zhang, and
  Zhou}]{yan2020prophetnet}
Yu~Yan, Weizhen Qi, Yeyun Gong, Dayiheng Liu, Nan Duan, Jiusheng Chen, Ruofei
  Zhang, and Ming Zhou. 2020.
\newblock Prophetnet: Predicting future n-gram for sequence-to-sequence
  pre-training.
\newblock \emph{arXiv preprint arXiv:2001.04063}.

\bibitem[{Yang et~al.(2019{\natexlab{a}})Yang, Zhang, Tar, and
  Baldridge}]{yang2019paws}
Yinfei Yang, Yuan Zhang, Chris Tar, and Jason Baldridge. 2019{\natexlab{a}}.
\newblock Paws-x: A cross-lingual adversarial dataset for paraphrase
  identification.
\newblock \emph{arXiv preprint arXiv:1908.11828}.

\bibitem[{Yang et~al.(2019{\natexlab{b}})Yang, Dai, Yang, Carbonell,
  Salakhutdinov, and Le}]{xlnet}
Zhilin Yang, Zihang Dai, Yiming Yang, Jaime Carbonell, Ruslan Salakhutdinov,
  and Quoc~V. Le. 2019{\natexlab{b}}.
\newblock Xlnet: Generalized autoregressive pretraining for language
  understanding.
\newblock \emph{NeurIPS}.

\bibitem[{Zeman et~al.(2019)Zeman, Nivre, Abrams, Aepli, Agi{\'c}, Ahrenberg,
  Aleksandravi{\v c}i{\=u}t{\.e}, Antonsen, Aplonova, Aranzabe, Arutie,
  Asahara, Ateyah, Attia, Atutxa, Augustinus, Badmaeva, Ballesteros, Banerjee,
  Bank, Barbu~Mititelu, Basmov, Batchelor, Bauer, Bellato, Bengoetxea, Berzak,
  Bhat, Bhat, Biagetti, Bick, Bielinskien{\.e}, Blokland, Bobicev, Boizou,
  Borges~V{\"o}lker, B{\"o}rstell, Bosco, Bouma, Bowman, Boyd, Brokait{\.e},
  Burchardt, Candito, Caron, Caron, Cavalcanti, Cebiro{\u g}lu~Eryi{\u g}it,
  Cecchini, Celano, {\v C}{\'e}pl{\"o}, Cetin, Chalub, Choi, Cho, Chun,
  Cignarella, Cinkov{\'a}, Collomb, {\c C}{\"o}ltekin, Connor, Courtin,
  Davidson, de~Marneffe, de~Paiva, de~Souza, Diaz~de Ilarraza, Dickerson,
  Dione, Dirix, Dobrovoljc, Dozat, Droganova, Dwivedi, Eckhoff, Eli, Elkahky,
  Ephrem, Erina, Erjavec, Etienne, Evelyn, Farkas, Fernandez~Alcalde, Foster,
  Freitas, Fujita, Gajdo{\v s}ov{\'a}, Galbraith, Garcia, G{\"a}rdenfors,
  Garza, Gerdes, Ginter, Goenaga, Gojenola, G{\"o}k{\i}rmak, Goldberg,
  G{\'o}mez~Guinovart, Gonz{\'a}lez~Saavedra, Grici{\=u}t{\.e}, Grioni, Gr{\=
  u}z{\={\i}}tis, Guillaume, Guillot-Barbance, Habash, Haji{\v c}, Haji{\v
  c}~jr., H{\"a}m{\"a}l{\"a}inen, H{\`a}~M{\~y}, Han, Harris, Haug, Heinecke,
  Hennig, Hladk{\'a}, Hlav{\'a}{\v c}ov{\'a}, Hociung, Hohle, Hwang, Ikeda,
  Ion, Irimia, Ishola, Jel{\'{\i}}nek, Johannsen, J{\o}rgensen, Juutinen, Ka{\c
  s}{\i}kara, Kaasen, Kabaeva, Kahane, Kanayama, Kanerva, Katz, Kayadelen,
  Kenney, Kettnerov{\'a}, Kirchner, Klementieva, K{\"o}hn, Kopacewicz, Kotsyba,
  Kovalevskait{\.e}, Krek, Kwak, Laippala, Lambertino, Lam, Lando, Larasati,
  Lavrentiev, Lee, L{\^e}~H{\`{\^o}}ng, Lenci, Lertpradit, Leung, Li, Li, Li,
  Lim, Liovina, Li, Ljube{\v s}i{\'c}, Loginova, Lyashevskaya, Lynn, Macketanz,
  Makazhanov, Mandl, Manning, Manurung, M{\u a}r{\u a}nduc, Mare{\v c}ek,
  Marheinecke, Mart{\'{\i}}nez~Alonso, Martins, Ma{\v s}ek, Matsumoto,
  {McDonald}, {McGuinness}, Mendon{\c c}a, Miekka, Misirpashayeva, Missil{\"a},
  Mititelu, Mitrofan, Miyao, Montemagni, More, Moreno~Romero, Mori, Morioka,
  Mori, Moro, Mortensen, Moskalevskyi, Muischnek, Munro, Murawaki,
  M{\"u}{\"u}risep, Nainwani, Navarro~Hor{\~n}iacek, Nedoluzhko, Ne{\v
  s}pore-B{\=e}rzkalne, Nguy{\~{\^e}}n~Th{\d i}, Nguy{\~{\^e}}n Th{\d i}~Minh,
  Nikaido, Nikolaev, Nitisaroj, Nurmi, Ojala, Ojha, Ol{\'u}{\`o}kun, Omura,
  Osenova, {\"O}stling, {\O}vrelid, Partanen, Pascual, Passarotti, Patejuk,
  Paulino-Passos, Peljak-{\L}api{\'n}ska, Peng, Perez, Perrier, Petrova,
  Petrov, Phelan, Piitulainen, Pirinen, Pitler, Plank, Poibeau, Ponomareva,
  Popel, Pretkalni{\c n}a, Pr{\'e}vost, Prokopidis, Przepi{\'o}rkowski,
  Puolakainen, Pyysalo, Qi, R{\"a}{\"a}bis, Rademaker, Ramasamy, Rama, Ramisch,
  Ravishankar, Real, Reddy, Rehm, Riabov, Rie{\ss}ler, Rimkut{\.e}, Rinaldi,
  Rituma, Rocha, Romanenko, Rosa, Rovati, Roșca, Rudina, Rueter, Sadde, Sagot,
  Saleh, Salomoni, Samard{\v z}i{\'c}, Samson, Sanguinetti, S{\"a}rg,
  Saul{\={\i}}te, Sawanakunanon, Schneider, Schuster, Seddah, Seeker, Seraji,
  Shen, Shimada, Shirasu, Shohibussirri, Sichinava, Silveira, Silveira, Simi,
  Simionescu, Simk{\'o}, {\v S}imkov{\'a}, Simov, Smith, Soares-Bastos,
  Spadine, Stella, Straka, Strnadov{\'a}, Suhr, Sulubacak, Suzuki,
  Sz{\'a}nt{\'o}, Taji, Takahashi, Tamburini, Tanaka, Tellier, Thomas, Torga,
  Trosterud, Trukhina, Tsarfaty, Tyers, Uematsu, Ure{\v s}ov{\'a}, Uria,
  Uszkoreit, Utka, Vajjala, van Niekerk, van Noord, Varga, Villemonte de~la
  Clergerie, Vincze, Wallin, Walsh, Wang, Washington, Wendt, Williams,
  Wir{\'e}n, Wittern, Woldemariam, Wong, Wr{\'o}blewska, Yako, Yamazaki, Yan,
  Yasuoka, Yavrumyan, Yu, {\v Z}abokrtsk{\'y}, Zeldes, Zhang, and Zhu}]{ud2.5}
Daniel Zeman, Joakim Nivre, Mitchell Abrams, No{\"e}mi Aepli, {\v Z}eljko
  Agi{\'c}, Lars Ahrenberg, Gabriel{\.e} Aleksandravi{\v c}i{\=u}t{\.e}, Lene
  Antonsen, Katya Aplonova, Maria~Jesus Aranzabe, Gashaw Arutie, Masayuki
  Asahara, Luma Ateyah, Mohammed Attia, Aitziber Atutxa, Liesbeth Augustinus,
  Elena Badmaeva, Miguel Ballesteros, Esha Banerjee, Sebastian Bank, Verginica
  Barbu~Mititelu, Victoria Basmov, Colin Batchelor, John Bauer, Sandra Bellato,
  Kepa Bengoetxea, Yevgeni Berzak, Irshad~Ahmad Bhat, Riyaz~Ahmad Bhat, Erica
  Biagetti, Eckhard Bick, Agn{\.e} Bielinskien{\.e}, Rogier Blokland, Victoria
  Bobicev, Lo{\"{\i}}c Boizou, Emanuel Borges~V{\"o}lker, Carl B{\"o}rstell,
  Cristina Bosco, Gosse Bouma, Sam Bowman, Adriane Boyd, Kristina Brokait{\.e},
  Aljoscha Burchardt, Marie Candito, Bernard Caron, Gauthier Caron, Tatiana
  Cavalcanti, G{\"u}l{\c s}en Cebiro{\u g}lu~Eryi{\u g}it, Flavio~Massimiliano
  Cecchini, Giuseppe G.~A. Celano, Slavom{\'{\i}}r {\v C}{\'e}pl{\"o}, Savas
  Cetin, Fabricio Chalub, Jinho Choi, Yongseok Cho, Jayeol Chun, Alessandra~T.
  Cignarella, Silvie Cinkov{\'a}, Aur{\'e}lie Collomb, {\c C}a{\u g}r{\i} {\c
  C}{\"o}ltekin, Miriam Connor, Marine Courtin, Elizabeth Davidson,
  Marie-Catherine de~Marneffe, Valeria de~Paiva, Elvis de~Souza, Arantza
  Diaz~de Ilarraza, Carly Dickerson, Bamba Dione, Peter Dirix, Kaja Dobrovoljc,
  Timothy Dozat, Kira Droganova, Puneet Dwivedi, Hanne Eckhoff, Marhaba Eli,
  Ali Elkahky, Binyam Ephrem, Olga Erina, Toma{\v z} Erjavec, Aline Etienne,
  Wograine Evelyn, Rich{\'a}rd Farkas, Hector Fernandez~Alcalde, Jennifer
  Foster, Cl{\'a}udia Freitas, Kazunori Fujita, Katar{\'{\i}}na Gajdo{\v
  s}ov{\'a}, Daniel Galbraith, Marcos Garcia, Moa G{\"a}rdenfors, Sebastian
  Garza, Kim Gerdes, Filip Ginter, Iakes Goenaga, Koldo Gojenola, Memduh
  G{\"o}k{\i}rmak, Yoav Goldberg, Xavier G{\'o}mez~Guinovart, Berta
  Gonz{\'a}lez~Saavedra, Bernadeta Grici{\=u}t{\.e}, Matias Grioni, Normunds
  Gr{\= u}z{\={\i}}tis, Bruno Guillaume, C{\'e}line Guillot-Barbance, Nizar
  Habash, Jan Haji{\v c}, Jan Haji{\v c}~jr., Mika H{\"a}m{\"a}l{\"a}inen, Linh
  H{\`a}~M{\~y}, Na-Rae Han, Kim Harris, Dag Haug, Johannes Heinecke, Felix
  Hennig, Barbora Hladk{\'a}, Jaroslava Hlav{\'a}{\v c}ov{\'a}, Florinel
  Hociung, Petter Hohle, Jena Hwang, Takumi Ikeda, Radu Ion, Elena Irimia, {\d
  O}l{\'a}j{\'{\i}}d{\'e} Ishola, Tom{\'a}{\v s} Jel{\'{\i}}nek, Anders
  Johannsen, Fredrik J{\o}rgensen, Markus Juutinen, H{\"u}ner Ka{\c s}{\i}kara,
  Andre Kaasen, Nadezhda Kabaeva, Sylvain Kahane, Hiroshi Kanayama, Jenna
  Kanerva, Boris Katz, Tolga Kayadelen, Jessica Kenney, V{\'a}clava
  Kettnerov{\'a}, Jesse Kirchner, Elena Klementieva, Arne K{\"o}hn, Kamil
  Kopacewicz, Natalia Kotsyba, Jolanta Kovalevskait{\.e}, Simon Krek, Sookyoung
  Kwak, Veronika Laippala, Lorenzo Lambertino, Lucia Lam, Tatiana Lando,
  Septina~Dian Larasati, Alexei Lavrentiev, John Lee, Phuong
  L{\^e}~H{\`{\^o}}ng, Alessandro Lenci, Saran Lertpradit, Herman Leung,
  Cheuk~Ying Li, Josie Li, Keying Li, {KyungTae} Lim, Maria Liovina, Yuan Li,
  Nikola Ljube{\v s}i{\'c}, Olga Loginova, Olga Lyashevskaya, Teresa Lynn,
  Vivien Macketanz, Aibek Makazhanov, Michael Mandl, Christopher Manning, Ruli
  Manurung, C{\u a}t{\u a}lina M{\u a}r{\u a}nduc, David Mare{\v c}ek, Katrin
  Marheinecke, H{\'e}ctor Mart{\'{\i}}nez~Alonso, Andr{\'e} Martins, Jan Ma{\v
  s}ek, Yuji Matsumoto, Ryan {McDonald}, Sarah {McGuinness}, Gustavo Mendon{\c
  c}a, Niko Miekka, Margarita Misirpashayeva, Anna Missil{\"a}, C{\u a}t{\u
  a}lin Mititelu, Maria Mitrofan, Yusuke Miyao, Simonetta Montemagni, Amir
  More, Laura Moreno~Romero, Keiko~Sophie Mori, Tomohiko Morioka, Shinsuke
  Mori, Shigeki Moro, Bjartur Mortensen, Bohdan Moskalevskyi, Kadri Muischnek,
  Robert Munro, Yugo Murawaki, Kaili M{\"u}{\"u}risep, Pinkey Nainwani,
  Juan~Ignacio Navarro~Hor{\~n}iacek, Anna Nedoluzhko, Gunta Ne{\v
  s}pore-B{\=e}rzkalne, Luong Nguy{\~{\^e}}n~Th{\d i}, Huy{\`{\^e}}n
  Nguy{\~{\^e}}n Th{\d i}~Minh, Yoshihiro Nikaido, Vitaly Nikolaev, Rattima
  Nitisaroj, Hanna Nurmi, Stina Ojala, Atul~Kr. Ojha, Ad{\'e}day{\d o}
  Ol{\'u}{\`o}kun, Mai Omura, Petya Osenova, Robert {\"O}stling, Lilja
  {\O}vrelid, Niko Partanen, Elena Pascual, Marco Passarotti, Agnieszka
  Patejuk, Guilherme Paulino-Passos, Angelika Peljak-{\L}api{\'n}ska, Siyao
  Peng, Cenel-Augusto Perez, Guy Perrier, Daria Petrova, Slav Petrov, Jason
  Phelan, Jussi Piitulainen, Tommi~A Pirinen, Emily Pitler, Barbara Plank,
  Thierry Poibeau, Larisa Ponomareva, Martin Popel, Lauma Pretkalni{\c n}a,
  Sophie Pr{\'e}vost, Prokopis Prokopidis, Adam Przepi{\'o}rkowski, Tiina
  Puolakainen, Sampo Pyysalo, Peng Qi, Andriela R{\"a}{\"a}bis, Alexandre
  Rademaker, Loganathan Ramasamy, Taraka Rama, Carlos Ramisch, Vinit
  Ravishankar, Livy Real, Siva Reddy, Georg Rehm, Ivan Riabov, Michael
  Rie{\ss}ler, Erika Rimkut{\.e}, Larissa Rinaldi, Laura Rituma, Luisa Rocha,
  Mykhailo Romanenko, Rudolf Rosa, Davide Rovati, Valentin Roșca, Olga Rudina,
  Jack Rueter, Shoval Sadde, Beno{\^{\i}}t Sagot, Shadi Saleh, Alessio
  Salomoni, Tanja Samard{\v z}i{\'c}, Stephanie Samson, Manuela Sanguinetti,
  Dage S{\"a}rg, Baiba Saul{\={\i}}te, Yanin Sawanakunanon, Nathan Schneider,
  Sebastian Schuster, Djam{\'e} Seddah, Wolfgang Seeker, Mojgan Seraji,
  Mo~Shen, Atsuko Shimada, Hiroyuki Shirasu, Muh Shohibussirri, Dmitry
  Sichinava, Aline Silveira, Natalia Silveira, Maria Simi, Radu Simionescu,
  Katalin Simk{\'o}, M{\'a}ria {\v S}imkov{\'a}, Kiril Simov, Aaron Smith,
  Isabela Soares-Bastos, Carolyn Spadine, Antonio Stella, Milan Straka, Jana
  Strnadov{\'a}, Alane Suhr, Umut Sulubacak, Shingo Suzuki, Zsolt
  Sz{\'a}nt{\'o}, Dima Taji, Yuta Takahashi, Fabio Tamburini, Takaaki Tanaka,
  Isabelle Tellier, Guillaume Thomas, Liisi Torga, Trond Trosterud, Anna
  Trukhina, Reut Tsarfaty, Francis Tyers, Sumire Uematsu, Zde{\v n}ka Ure{\v
  s}ov{\'a}, Larraitz Uria, Hans Uszkoreit, Andrius Utka, Sowmya Vajjala,
  Daniel van Niekerk, Gertjan van Noord, Viktor Varga, Eric Villemonte de~la
  Clergerie, Veronika Vincze, Lars Wallin, Abigail Walsh, Jing~Xian Wang,
  Jonathan~North Washington, Maximilan Wendt, Seyi Williams, Mats Wir{\'e}n,
  Christian Wittern, Tsegay Woldemariam, Tak-sum Wong, Alina Wr{\'o}blewska,
  Mary Yako, Naoki Yamazaki, Chunxiao Yan, Koichi Yasuoka, Marat~M. Yavrumyan,
  Zhuoran Yu, Zden{\v e}k {\v Z}abokrtsk{\'y}, Amir Zeldes, Manying Zhang, and
  Hanzhi Zhu. 2019.
\newblock \href {http://hdl.handle.net/11234/1-3105} {Universal dependencies
  2.5}.
\newblock {LINDAT}/{CLARIAH}-{CZ} digital library at the Institute of Formal
  and Applied Linguistics ({{\'U}FAL}), Faculty of Mathematics and Physics,
  Charles University.

\bibitem[{Zhang et~al.(2019)Zhang, Baldridge, and He}]{zhang2019paws}
Yuan Zhang, Jason Baldridge, and Luheng He. 2019.
\newblock Paws: Paraphrase adversaries from word scrambling.
\newblock \emph{arXiv preprint arXiv:1904.01130}.

\bibitem[{Zhou et~al.(2020)Zhou, Palangi, Zhang, Hu, Corso, and Gao}]{vlp}
Luowei Zhou, Hamid Palangi, Lei Zhang, Houdong Hu, Jason Corso, and Jianfeng
  Gao. 2020.
\newblock Unified vision-language pre-training for image captioning and vqa.
\newblock \emph{AAAI}.

\end{thebibliography}
\bibliographystyle{style/acl_natbib}
\end{document}